\documentclass[twoside,11pt]{article}

\usepackage{blindtext}

\usepackage{longtable}

\usepackage{booktabs}
\usepackage{comment}
\usepackage{graphicx}
\usepackage{subcaption}

\usepackage{jmlr2e}

\usepackage{lastpage}
\jmlrheading{23}{2022}{1-\pageref{LastPage}}{1/21; Revised 5/22}{9/22}{21-0000}{Balaji Ganesan et al}

\ShortHeadings{xLP: Explainable Link Prediction for Master Data Management}{xLP: Explainable Link Prediction for Master Data Management}
\firstpageno{1}

\begin{document}

\title{xLP: Explainable Link Prediction \\for Master Data Management}

\author{\name Balaji Ganesan \email bganesa1@in.ibm.com \\
\name{Sumit Bhatia} \email{sumitbhatia@in.ibm.com}\\
\name{Hima Patel} \email{himapatel@in.ibm.com}\\
\name{Sameep Mehta} \email{sameepmehta@in.ibm.com}\\
\addr IBM Research, India
\AND
\name{Matheen {Ahmed Pasha}}\email{matpasha@in.ibm.com} \\
\name{Srinivasa Parkala} \email{shsriniv@in.ibm.com} \\
\name{Neeraj {R Singh}} \email{sneeraj@in.ibm.com} \\
\name{Gayatri Mishra}\email{gayamish@in.ibm.com} \\
\name{Somashekhar Naganna} \email{soma.shekar@in.ibm.com} \\
\addr IBM Data and AI, India
}
\editor{}

\maketitle

\begin{abstract}
Explaining neural model predictions to users requires creativity. Especially in enterprise applications, where there are costs associated with users' time, and their trust in the model predictions is critical for adoption. For link prediction in master data management, we have built a number of explainability solutions drawing from research in interpretability, fact verification, path ranking, neuro-symbolic reasoning and self-explaining AI. In this demo, we present explanations for link prediction in a creative way, to allow users to choose explanations they are more comfortable with.
\end{abstract}

\begin{keywords}
Link Prediction, Graph Neural Networks, Explainability
\end{keywords}

\section{Introduction}
\label{sec:intro}

Link prediction using Graph Neural Networks (GNN) is a well researched problem. However, there remain a number of challenges in using GNN models in enterprise applications. Especially when the nodes of the graphs are people or organizations who are customers of an enterprise, the links are real world relationships, and the applications are sensitive like customer due diligence, anti-money laundering, and law enforcement. In \cite{ganesan2020link}, we discussed ethical considerations and practical insights from developing a link prediction solution for IBM's Infosphere Master Data Management (MDM) product.

MDM users include data engineers who work on ETL, data stewards who help customers with managing Master Data, and data scientists who use MDM for analytics. Their expectations from a explainability solution are different. While some might be interested in model interpretability, others might want a quick and easy to understand solution, to get the job done. In this demo, we compare three explainability techniques for Link Prediction.
\begin{itemize}
    \item Interpretable models that approximate neural model predictions.
    \item Link verification where external information corroborates the predictions.
    \item Path ranking algorithms that were earlier used for error detection.
\end{itemize}

While evaluating the explanations, the hypothesis is that users who have seen neural model predictions and explanations, should later be able to \textit{guess} what the neural model will predict on unseen data.

One approach to selecting an explanation from the above choices, involves pre-selecting some explanations suitable for the audience. Few such demos, where we tailored the explanations to the users, led us to design a case-study using an interactive user interface. Our goal here is to understand user preferences towards different types of explanations, their ability to understand the models, and also validate what constitutes an human understandable explanation.

\begin{figure}[htbp]
    \centering
    \includegraphics[width=\linewidth]{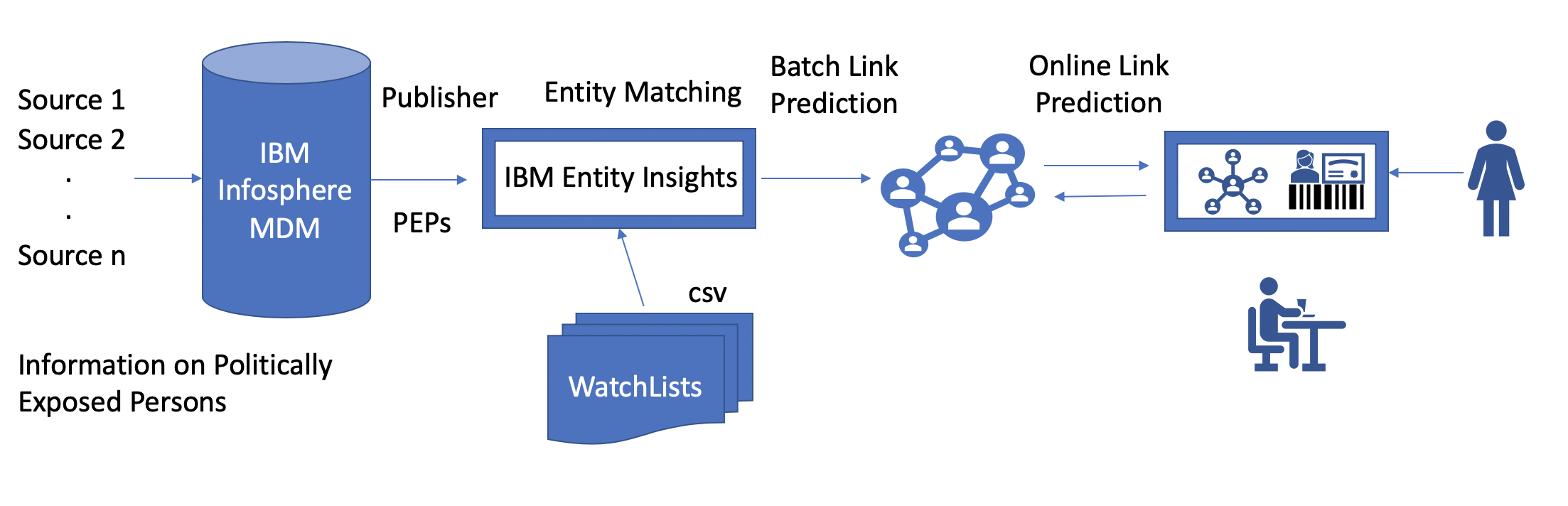}
    \label{fig:watchlist_link_prediction}
    \caption{Predicting links to watch-list nodes. A list of COVID19 persons can be uploaded as watch-list and their links to people in a master data could be predicted.}
\end{figure}

Master Data Management includes tasks like entity matching, link prediction, non-obvious relation extraction among other things. Typically the entities in master data are people, organizations and location. Though products, events and other entities can be stored too. While this master data is predominantly stored as relational data, graph stores are widely used for visualization and analytics. In this work, we focus on link prediction. Graphs in MDM could be considered as property graphs which differ from knowledge graphs and social networks. In particular, MDM graphs often are homogenous attributed graphs where one of people, organization, or location constitute nodes in the graph. Other entities like numerical ids, demographic information, business terms constitute the attributes of the nodes. Links between people or organizations in the real world constitute the edges of the graph. The type of people to people relations form the link type and details of the relationship like duration constitute the edge properties.

Link Prediction on people graphs presents a unique set of challenges in terms of model performance, data availability, fairness, privacy, and data protection. Further, Link Prediction among people has a number of societal implications irrespective of the use-cases. While link prediction of financial fraud detection, law enforcement, and advertisement go through ethical scrutiny, the recent use of social networks for political targeting and job opportunities also require ethical awareness on the part of the model developers and system designers.

Watchlist is a use-case typical in Master Data Management. Given a set of nodes in a graph, the task is to find links to Watchlist nodes from other nodes of the graph. For example, we can assume people who have tested positive for COVID19 as people on a Watchlist. We may want to find people who may have potentially come in contact with them, commonly known as contact tracing. This can however be a time consuming and potentially controversial process that impacts privacy. We need informed consent of people for this contact tracing. In this work we discuss link prediction on property graphs involving people, their specific dataset requirements, ethical considerations, and practical insights in designing the infrastructure for industrial scale deployment.

\section{Related Work}

\cite{schlichtkrull2018modeling} showed that a Relational Graph Convolutional Network can outperform direct optimization of the factorization (ex: DistMult). They used an autoencoder model consisting of an encoder – an R-GCN producing latent feature representations of entities and a decoder – a tensor factorization model exploiting these representations to predict labeled edges.

\cite{hamilton2017inductive} introduced GraphSAGE (SAmple and AggreGatE) an inductive framework that leverages node feature information (ex: text attributes, node degrees) to efficiently generate node embeddings for previously unseen data or entirely new (sub)graphs. In this inductive framework, we learn a function that generates embeddings by sampling and aggregating features from a node’s local neighborhood. \cite{you2019position} proposed a Position Aware Graph Neural Network that significantly improves performance on the Link Prediction task over the Graph Convolutional Networks.

\cite{guan2019link} introduced the WikiPeople dataset based on Wikidata. However, Wikidata does not have contact details and can be incomplete like the DBPedia (UDBMS) dataset. \cite{dasgupta2018hyte} also introduced a dataset based on Wikidata. Again, for afore mentioned reasons, it’s not different from DBPedia dataset. \cite{Zitnik2017} introduced the Protein-Protein Interaction dataset which has been used in a number of recent works in graph neural networks. \cite{kipf2017semisupervised} presented Graph Convolutional networks. \cite{hamilton2017inductive} introduced the idea of inductive representation of nodes in a graph. \cite{you2019position} added anchor nodes to improve the representation of nodes in the graph.

\cite{oberhofer2014beyond} introduced many aspects of entity resolution using a probabilistic matching engine. DeepMatcher \cite{mudgal2018deep} presents a neural model for Entity Matching. \cite{konda2019executing} presented an end-to-end use case for Entity Matching. \cite{mullerintegrated} described an integrated neural model to find non-obvious relations. The Protein-Protein Interaction (PPI) dataset introduced by \cite{Zitnik2017} consists of 24 human tissues and hence has 24 subgraphs of roughly 2400 nodes each and their edges. Having similar subgraphs helps to average the performance of the model across subgraphs. Open Graph Benchmark \cite{ogb_benchmark} initiative by SNAP Group at Stanford is trying to come up with large benchmark datasets for research in Graph Neural Models.

Much of the recent work on explanations are based on post-hoc models that try to approximate the prediction of complex models using interpretable models. \cite{vannur2020data} present post-hoc explanations of the links predicted by a Graph Neural Network by treating it as a classification problem. They present explanations using LIME \cite{ribeiro2016should} and SHAP \cite{lundberg2017unified}.

Over the years, Attention has been understood to provide an important way to explain the workings of neural models, particularly in the field of NLP. \cite{jain2019attention} challenged this understanding by showing that the assumptions for accepting Attention as explanation do not hold. \cite{wiegreffe2019attention} however argued that Attention also contributes to explainability. More recently, \cite{agarwal2020neural} introduced Neural Additive Models which learn a model for each feature to increase the interpretability. In this work, we focus on solutions that are more suited to Graph Neural Models and in particular, explaining Link Prediction models.

\section{Link Prediction}
\label{experiments}

We begin by explaining our experimental setup. We use the framework provided by \cite{you2019position}, which in turn uses pytorch \cite{paszke2019pytorch} and more specifically pytorch geometric \cite{fey2019fast}. One of the pre-processing steps we do is finding the all pairs shortest paths calculation using appropriate approximations. As we'll see in Section \ref{explainability}, this pre-processing step comes in handy to explain the links as well.

We train models using GCN and PGNN on the UDBMS Dataset and the MDM Bootcamp Dataset that is based on the wikipedia. Following the procedure in \cite{you2019position}, we choose only connected components with atleast 10 nodes for our experiments. A positive sample is created by randomly choosing 10\% of the links. For the negative sample, we use one of the nodes involved in the positive samples and pick a random unconnected node as the other node. The number of negative samples is same as that of the positive samples. We discuss hard negative samples in our future work section. Our batch size is typically 8 subgraphs and for PGNN, we use 64 anchor nodes.

\begin{table}[!htb]
    \begin{center}
    \begin{tabular}{cccc}
    \hline
    \addlinespace
    \textbf{Dataset} & \textbf{Model} & \textbf{ROC AUC} & \textbf{Std. Dev.} \\
        \addlinespace
    \hline
    \addlinespace
    {\textbf{UDBMS}}  &   \textbf{GCN} & 0.4689 & 0.0280   \\
                    &   \textbf{P-GNN} & 0.6456 & 0.0185   \\
                        \cline{2-4}
    \addlinespace
    {\textbf{MDM Dataset}}  &   \textbf{GCN} & 0.4047  & 0.09184  \\
                    &   \textbf{P-GNN} & 0.6473 & 0.02116   \\
                                            \cline{2-4}
    \hline
    \end{tabular}
    \caption{Comparison of Link Prediction performance on different People Datasets}
    \label{tab:comparison}
    \end{center}
\end{table}

As can be seen in Table \ref{tab:comparison}, P-GNN outperforms GCN in both our datasets. The performance obtained by the better model is also less than what we desire. There are a number of options available to try and improve the performance of the models. However, in this work, we want to focus on how such a system can be deployed in production under the professional oversight of Data Stewards.

\begin{figure*}[htb]
    \centering
    \includegraphics[width=\linewidth]{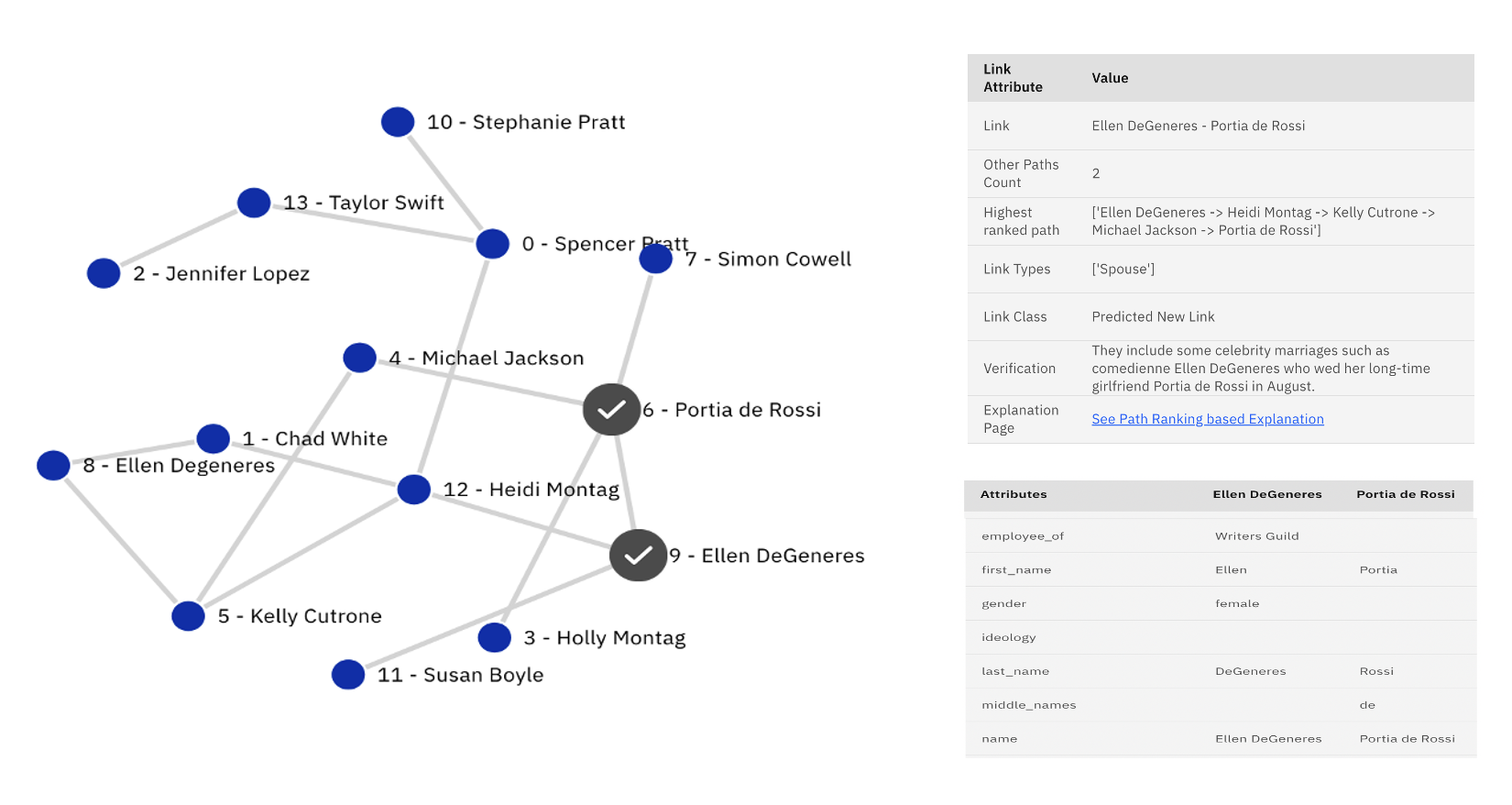}
    \caption{xLP: Explainable Link Prediction. The above image shows a subgraph with links predicted. On the right we see verification text for the predicted link and a comparison of the nodes involved in the link.}
    \label{fig:xlp}
\end{figure*}

\subsection{Explainability}
\label{explainability}

In typical enterprise deployments, it is desired that the models provide explanations so that the Data Stewards can understand why the model is predicting a link. While Data Stewards can also use the system to detect errors, we want the Data Stewards to draw valuable insights into the model's learning process and also understand it's limitations. In this section, we discuss several explainability techniques that could be used for explaining the links predicted by the Graph Neural Network models in Section \ref{experiments}. While a number of above such solutions for explaining neural models exist, our focus is been on human understandable explanations for graph neural networks. We explored different techniques in this area and categorized them into three.

\subsubsection*{Verification}
Verification is the proof for the predicted link. This can be used to understand links already present in the graph and also try to validate new links predicted by our models. Some of the earlier error detection work involved external information to verify the predicted link. \cite{thorne2018fever} describes the FEVER shared task for facts extraction and verification.

\subsubsection*{Interpretation}
\cite{ribeiro2018anchors} introduced the idea of Anchors as Explanations, which builds upon their earlier LIME solution \cite{ribeiro2016should}. The key idea here is to show only few important features (anchors) rather than showing the pros and cons of several or all features. \cite{huang2020graphlime} introduced GraphLIME, which as the name indicates is a version of LIME for Graph Neural Networks. They compared their solution GNN Explainer \cite{ying2019gnnexplainer} which follows a subgraph approach to explain predicted links.

\subsubsection*{Explanation}

\cite{elton2020self} presents the Self-Explainable AI idea where a single model is used to both predict and provide parameters that can be used to explain the prediction. This is similar to a multi-task learning setting. Mutual Information is used to measure the faithfulness of the explanation model to the diagnostic model. \cite{sun2019infograph} also uses Mutual Information to measure the closeness of supervised and unsupervised models.

\section{Implementation}

We used GNN Explainer \cite{ying2019gnnexplainer} as the baseline explanation which we want to make more human understandable using the 3 techniques discussed in this work.

\begin{figure}[htbp]
    \centering
    \includegraphics[width=\linewidth]{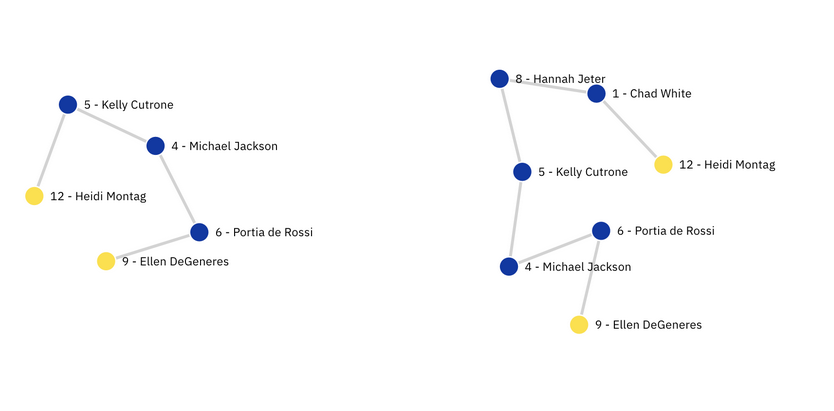}
    \caption{Path Ranking based explanation for Link Prediction. Data Stewards can explore the already existing paths (if any) and understand the predicted link. They can also provide feedback.}
    \label{fig:pbe}
\end{figure}

\subsection{Link Verification with External Information}

One of the desirable properties of the dataset that we set out to create is the ability to verify the links present in the dataset. \cite{thorne2018fever} introduced the FEVER dataset for fact verification. We follow a similar method in our work to generate verifiable text that a human annotator can use to decide if the predicted link is valid or otherwise.

While in this dataset, we use text from unstructured pages, we can also include links from other sources namely organization charts, logs, emails and other enterprise documents to generate such verification text. The verifiable text for our example is shown in Figure \ref{fig:xlp}. For links that are present in training data, it'll be easy to fetch the verifying text. During inference on unseen data, we could using an information retrieval system, typically a search index to query for text and other details that mention the entities that are predicted to be linked by the Graph Neural Network.

\begin{figure}[htbp]
    \centering
    \includegraphics[width=\linewidth]{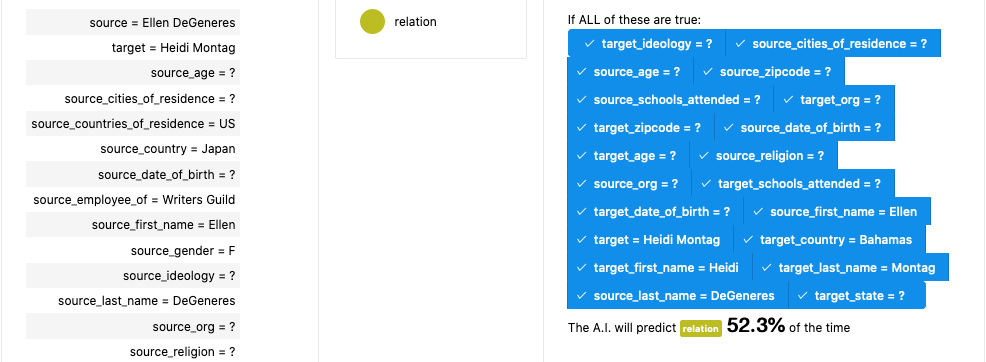}
    \caption{Anchors Explanation}
    \label{fig:anchors_explanations}
\end{figure}

\subsection{Anchors for Link Explanation}

We train a simple classifier to predict if a source and target nodes are linked or otherwise. This classifier is trained based on the output of the GNN model described in Section \ref{experiments}. Compared to the GNN model, this post hoc classification model is considered to be more interpretable.

In the above explanation, the classifier model is predicting a link between Ellen DeGeneres and Heidi Montag just like the original GNN model. Anchors model then explains that these two nodes will have a link 52.3\% of the time, if the conditions shown in Figure \ref{fig:anchors_explanations} hold. This result suffers from loss of information between the GNN model and the post-hoc interpretable model. Our classifier above is rudimentary and with more feature engineering, we could make the classifier model closer to the GNN model. However, even with reduced accuracy, providing a summary of the conditions under which a model predicts a link is more intuitive than the more complex graph explainability solutions.

\subsection{Path Ranking based Link Explanation}

Our next explainability solution is inspired by ideas in Error detection in Knowledge Graphs. In particular, we use the PaTyBRED approach described in \cite{meloautomatic}. Path Ranking Algorithm (PRA) introduced by \cite{lao2011random} was based on the probability of ending up at an Object o, if we perform random walks on a graph starting at Subject s and Relation r. \cite{gardner-mitchell-2015-efficient} replaced the path probability with binary values. PaTyBRED further simplified the process by using a Random Forest instead of logistic regression and introducing a k-selection method.

As shown in Figure \ref{fig:pbe}, we use the ranking function in the PaTyBRED algorithm to find a path among multiple already existing paths between two nodes. The idea here is that the information contained in that path is more useful than other paths to understand the predicted link. This idea is somewhat similar to the GNN Explainer idea of exploring the subgraph around the predicted link, except that we prefer to use an independent algorithm to rank all the paths rather than subgraph around the two nodes.

\subsection{Evaluation}

We then evaluated each of these explainability techniques in the form of a case study. We presented 100 samples of predicted links, the ground truth, and each of the 3 explanations along with GNN explainer output. We measured how often human annotators agree that the explanation is accurate for the predicted link. We did not evaluate based on which explanation is better since we generally got subjective answers in the initial part of the case study. Annotators preferred having multiple explainability solutions, each of which might be performing better in certain aspects. So we asked the annotators how often they agreed with each of the methods. The results from this case study are as shown in Table \ref{tab:evaluation}.

\begin{table}[!htb]
    \begin{center}
    \begin{tabular}{lr}
    \hline
    \addlinespace
    \textbf{Explainability Technique} & \textbf{Agreement} \\
    \addlinespace
    \hline
     \addlinespace
    {\textbf{Link Verification}} &  81   \\
    {\textbf{Anchors based Explanation}}   & 56  \\
    {\textbf{Path Ranking based Explanation}}   & 63  \\
     \addlinespace
    \hline
    \end{tabular}
    \caption{Comparison of annotator agreement with explanations}
    \label{tab:evaluation}
    \end{center}
\end{table}

\section{Conclusion}
In this demo, we presented a solution for the Link Prediction task on property graphs for Master Data Management. We presented a number of motivating experiments and use-cases, and describe the special care we take to address fairness, data protection, privacy and AI ethics. We then presented three human understandable solutions to explain the links predicted by Graph Neural Networks, namely search based retrieval of verification text, anchors based, and path ranking based explanations.

\begin{acks}
We thank Scott Schumacher, Marcus Boon, Berthold Reinwald, Xiao Qin, Jim O'Neil, Martin Oberhofer, Lars Bremmer, Manfred Overs, and Philip Mueller.
\end{acks}

\vskip 0.2in
\bibliography{neurips2020}

\end{document}